\documentclass[10pt,letterpaper]{article}

\usepackage{cogsci}
\usepackage[format=plain,
            labelfont=it,
            textfont=it]{caption}

\usepackage[utf8]{inputenc}
\usepackage{graphicx}
\usepackage{amsmath,amsfonts,amsthm,color}
\usepackage{tabularx}
\usepackage{array}
\usepackage{setspace}
\usepackage{color}
\usepackage{listings}
\usepackage{multirow}
\usepackage{tabularx}
\usepackage{ifthen}
\usepackage[T1]{fontenc}
\usepackage{framed}
\usepackage{pdfpages}
\usepackage{longtable}
\usepackage{enumitem}
\usepackage{times}
\usepackage{todonotes}
\usepackage{placeins}

\usepackage[
    colorlinks,
    citecolor=blue,
    linkcolor=blue,
    urlcolor=blue,
]{hyperref}

\graphicspath{{thesis/introduction/}{thesis/agents/}{thesis/requirements/}{thesis/approaches/}{thesis/architecture/}{thesis/actions/}{thesis/evaluation/}{thesis/interfacedeployment/}{thesis/conclusion/}}

\definecolor{delim}{RGB}{20,105,176}
\definecolor{numb}{RGB}{106, 109, 32}
\definecolor{string}{rgb}{0.64,0.08,0.08}
\definecolor{gray45}{rgb}{0.2, 0.5, 0.478}
\lstdefinelanguage{json}{
	showspaces=false,
	showtabs=false,
        showstringspaces=false,
	breaklines=true,
	postbreak=\raisebox{0ex}[0ex][0ex]{\ensuremath{\color{gray45}\hookrightarrow\space}},
	breakatwhitespace=true,
	basicstyle=\ttfamily\footnotesize,
        columns=fullflexible,
	upquote=true,
	morestring=[b]",
	stringstyle=\color{string},
	literate=
	*{0}{{{\color{numb}0}}}{1}
	{1}{{{\color{numb}1}}}{1}
	{2}{{{\color{numb}2}}}{1}
	{3}{{{\color{numb}3}}}{1}
	{4}{{{\color{numb}4}}}{1}
	{5}{{{\color{numb}5}}}{1}
	{6}{{{\color{numb}6}}}{1}
	{7}{{{\color{numb}7}}}{1}
	{8}{{{\color{numb}8}}}{1}
	{9}{{{\color{numb}9}}}{1}
	{\{}{{{\color{delim}{\{}}}}{1}
	{\}}{{{\color{delim}{\}}}}}{1}
	{[}{{{\color{delim}{[}}}}{1}
	{]}{{{\color{delim}{]}}}}{1},
}

\lstdefinelanguage{prompt}{
	showspaces=false,
	showtabs=false,
	breaklines=true,
	postbreak=\raisebox{0ex}[0ex][0ex]{\ensuremath{\color{gray45}\hookrightarrow\space}},
	breakatwhitespace=true,
	basicstyle=\ttfamily\footnotesize,
        columns=fullflexible,
	upquote=true,
	stringstyle=\color{string},
}

\definecolor{deepblue}{rgb}{0,0,0.5}
\definecolor{deepred}{rgb}{0.6,0,0}
\definecolor{deepgreen}{rgb}{0,0.5,0}
\DeclareFixedFont{\ttb}{T1}{txtt}{bx}{n}{9} 
\DeclareFixedFont{\ttm}{T1}{txtt}{m}{n}{9}  
\newcommand\pythonstyle{\lstset{
language=Python,
basicstyle=\ttfamily\footnotesize,
breaklines=true,
columns=fullflexible,
morekeywords={self},              
keywordstyle=\ttb\color{deepblue},
emph={MyClass,__init__},          
emphstyle=\ttb\color{deepred},    
stringstyle=\color{deepgreen},
frame=tb,                         
showstringspaces=false
}}
\lstnewenvironment{python}[1][]
{
\pythonstyle
\lstset{#1}
}
{}
\newcommand\pythoninline[1]{{\pythonstyle\lstinline!#1!}}
\title{Intelligent Assistants for the Semiconductor Failure Analysis with \\ LLM-Based Planning Agents}
 
\author{{\bf Aline Dobrovsky} \\
  \textit{University of the Bundeswehr Munich, 85579 Neubiberg, Germany}  \\
  \textit{aline.dobrovsky@unibw.de}
  \AND {\bf Konstantin Schekotihin} \\
  \textit{University Klagenfurt, 9020 Klagenfurt, Austria}\\
  \textit{konstantin.schekotihin@aau.at}
  \AND {\bf Christian Burmer} \\
  \textit{Infineon Technologies AG, 85579 Neubiberg, Germany}\\
  \textit{christian.burmer@infineon.com}
}

\begin{document}

\maketitle

\begin{abstract}
Failure Analysis (FA) is a highly intricate and knowledge-intensive process. The integration of AI components within the computational infrastructure of FA labs has the potential to automate a variety of tasks, including the detection of non-conformities in images, the retrieval of analogous cases from diverse data sources, and the generation of reports from annotated images. However, as the number of deployed AI models increases, the challenge lies in orchestrating these components into cohesive and efficient workflows that seamlessly integrate with the FA process.

This paper investigates the design and implementation of an agentic AI system for semiconductor FA using a Large Language Model (LLM)-based Planning Agent (LPA). The LPA integrates LLMs with advanced planning capabilities and external tool utilization, allowing autonomous processing of complex queries, retrieval of relevant data from external systems, and generation of human-readable responses. The evaluation results demonstrate the agent's operational effectiveness and reliability in supporting FA tasks.
\end{abstract}


%
%

\section{Introduction} 
\label{section:introduction}

The semiconductor industry underpins modern technology, producing essential components for diverse applications. Ensuring device correct functionality is critical, making Failure Analysis (FA) a key process in semiconductor manufacturing. FA systematically investigates non-conformities to identify their root causes and suggest corrective actions, supporting risk assessment and quality assurance. FA engineers require extensive expertise in science, engineering, and analysis processes, navigating vast technical data, including test results, microscopy images, and historical cases. Traditional knowledge-management systems, like databases, wikis, or file shares, demand significant manual effort, underscoring the need for AI-driven solutions to efficiently retrieve, synthesize, and present relevant information, enhancing decision-making and productivity in FA.

The rapid advancements in AI and Natural Language Processing (NLP) have opened new possibilities for enhancing complex, data-driven tasks across various industries. Large Language Models (LLMs) have recently emerged as powerful tools, capable of understanding human language and generating human-like text. By accessing and using human knowledge, they have the potential to significantly improve decision-making processes even in highly specialized fields.
However, general-purpose LLMs lack the domain-specific knowledge required for meaningful responses in FA, including knowledge about specific analysis methods and laboratory equipment, and domain-specific LLMs, like SemiKong \cite{DBLP:journals/corr/abs-2411-13802}, are still premature. As a result, modern solutions utilize semantic search and Retrieval-Augmented Generation (RAG) \cite{DBLP:journals/corr/abs-2312-10997}, to enhance their capabilities by using various prompting techniques \cite{Saravia_Guide_2022}. Experiments in the FA domain already demonstrated the benefits of these methods in improving the efficiency of information retrieval and decision-making processes \cite{10691083}. Nevertheless, these approaches still rely on a predefined pipeline and lack the flexibility to adapt to the specific needs of users. They also require a significant amount of data to be encoded and stored in a vector database, which can be time-consuming and resource-intensive.

This paper therefore explores the novel approach of LLM-based agents, which integrates LLMs with planning capabilities, memory, and tool utilization, such as database information retrieval, search machines, and even AI models, like object detectors or image classifiers. The LLM serves as the ``brain'' of an agent, decomposing complex queries into multiple tasks and resolving them step-by-step through reasoning and autonomous tool use. 
In particular, the suggested LLM-based Planning Agent (LPA) utilizes LLM-based ReAct planning \cite{DBLP:conf/iclr/YaoZYDSN023} as the core component to implement an AI assistant. To answer complex and ambiguous natural language questions from users, the system automatically identifies required information sources, synthesizes retrieval requests, analyzes obtained information, and generates textual responses. 
To retrieve the required data, we equip the LPA with a number of tools available on the cloud infrastructure of an FA lab, such as programming interfaces allowing search over documents on shares, wikis, and job logging databases, or access to other AI components, like image classifiers.
The tests and evaluation demonstrated that the agent is capable of autonomous, robust operation in a productive environment. It successfully interprets complex technical queries and autonomously uses reasoning and external tools to produce a textual answer suitable for an FA engineer. These achievements illustrate the potential of the LPA concept in a practical use case, namely supporting FA.

The paper is organized as follows: After an introduction to LLMs and intelligent agents, we detail the technical design, development, and implementation of the LPA application. Next, we evaluate agent's performance in generating high-quality responses to FA queries. Finally, we conclude with key findings and recommendations for future research.
\section[Theoretical Framework of LLM-Based Agents]{LLM-based Planning Agents}
\label{section:agents}

LLMs represent a significant advancement in NLP, leveraging the transformer architecture \cite{vaswani2017attention} to process and predict text based on patterns learned from extensive corpora. These models, with billions of parameters, excel in language understanding and generation tasks. 
Key milestones in their development include bidirectional transformers for contextual encoding \cite{devlin_bert_2019}, encoder-decoder architecture for text-to-text tasks \cite{raffel2020exploring}, and GPT-3 autoregressive model  \cite{brown_language_2020}. 
Nevertheless, LLMs often show limited performance for specific domains, like semiconductor FA, where domain-specific knowledge is crucial for effective problem-solving.

To address this, several strategies have been developed to enhance LLMs' capabilities. 
\emph{Instruction fine-tuning} \cite{ziegler_fine-tuning_2020,ouyang_training_2022} further enables LLMs to follow natural language instructions and adapt for domain-specific tasks. Similarly, \emph{in-context learning} or \emph{prompt-based learning} becomes increasingly popular alternative to the computationally intensive fine-tuning \cite{liu_pre-train_2021}. These methods automatically augment user input with textual prompts, like in Retrieval-Augmented Generation (RAG) \cite{DBLP:conf/nips/LewisPPPKGKLYR020}, as well as specific techniques, such as few-shot or Chain-of-Thought (CoT) prompting \cite{wei_chain--thought_2022}. For example, CoT prompts decompose complex problems into intermediate steps, significantly improving performance on reasoning tasks. Zero-shot CoT variants, such as appending ``let's think step by step'' to prompts, have also shown efficacy \cite{kojima_large_2023} in various tasks. Factors like the number of reasoning steps in CoT prompts further influence effectiveness, with more steps benefiting complex tasks \cite{jin_impact_2024}. However, both fine-tuning and prompting are based on specific pre-programmed pipelines, which may not be flexible enough to adapt to the specific needs of users and require much maintenance efforts to reflect changes in the data and software landscape of an FA lab. 

To address this issue, recent advancements in AI have facilitated the convergence of the classical concept of intelligent agents \cite{wooldridge_intelligent_1995} with the transformative capabilities of LLMs. \emph{Intelligent agents} are fundamentally characterized by their ability to perceive their environment and respond in a manner that aligns with predefined objectives. In the FA context, goal-oriented agents are particularly relevant as they leverage search and planning mechanisms to reach desirable goals, e.g., interpreting the states of a sample and corresponding analysis tasks. The belief-desire-intention architecture further enhances this paradigm by incorporating the internal goals of an agent, such as assisting an AI engineer in solving a problem~\cite{goos_intelligent_2002}. 
%
However, the applicability of these solutions in mission-critical enterprise environments remains to be thoroughly validated. Key challenges include seamlessly integrating tool handling with LLMs, such as combining database queries with the natural language processing capabilities of LLMs, accurately assessing tool functionalities to prevent unintended consequences (e.g., accidental database modifications or email dispatches), and implementing robust mechanisms for error handling during tool execution failures~\cite{muthusamy-etal-2023-towards}.


In the literature, LLM-based Planning Agents (LPA), still do not have a common architecture and might include different sets of components \cite{mialon_augmented_2023,wang_survey_2024,xi_rise_2023,zhao_-depth_2023}. In this paper, we consider the following LPA core components: (1) \emph{Plan Generation}, where the agent decomposes high-level tasks into executable subtasks and formulates a step-by-step plan; (2) \emph{Action Matching}, which involves selecting appropriate external tools to execute the subtasks; (3) \emph{Memory}, enabling the agent to retain information across multiple interactions; and (4) \emph{Feedback}, allowing the agent to refine its plan based on internal reflection or external outcomes.

\begin{figure}[bt]\begin{center}
	\includegraphics[width=0.43\textwidth]{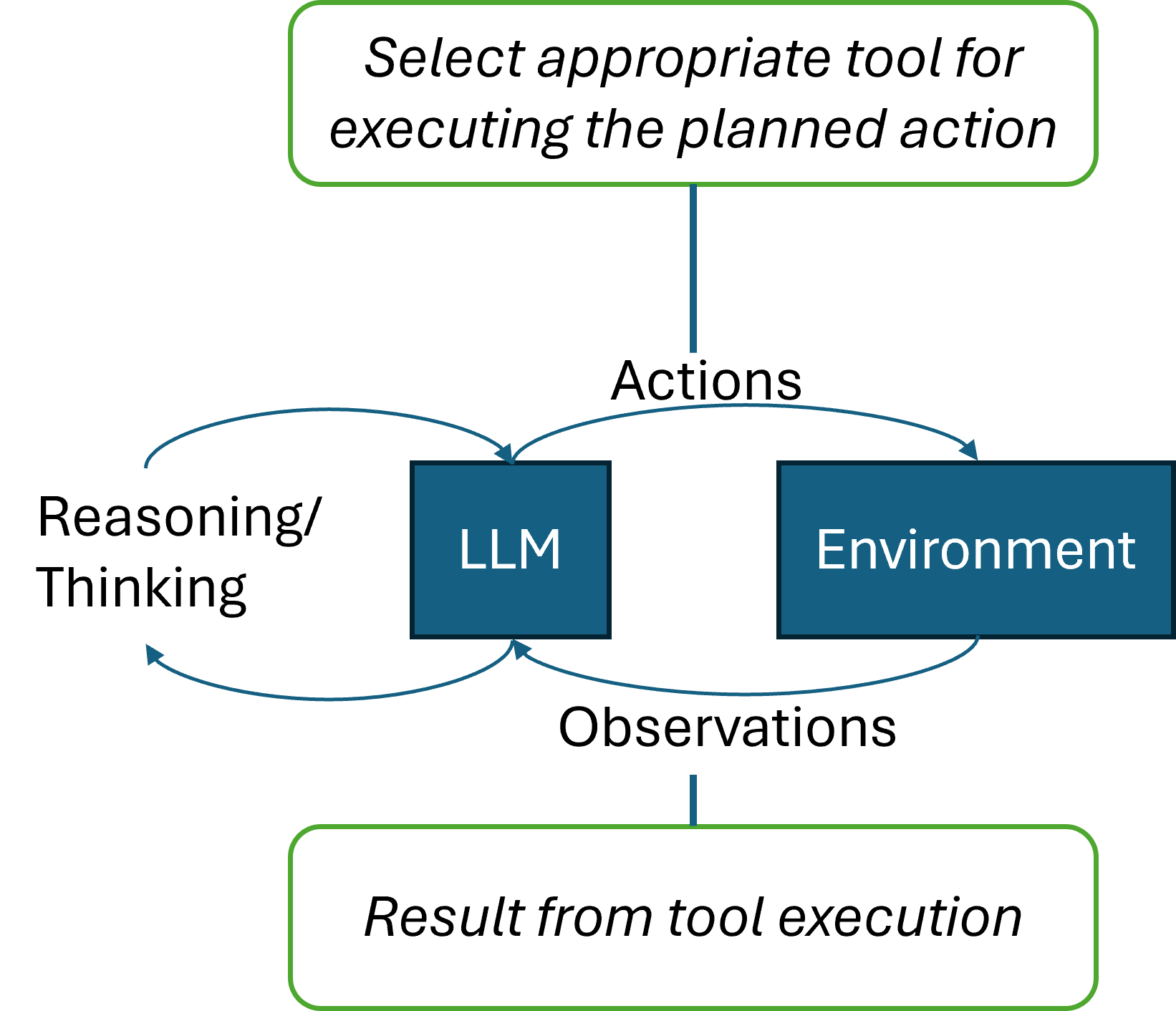}
	\caption{General schema of the ReAct approach \cite{DBLP:conf/iclr/YaoZYDSN023} to planning as iterative task generation}
	\label{fig:reactflow}
\end{center} \end{figure}

\begin{figure*}[bt]
	\centering
	\includegraphics[width=1.0\linewidth]{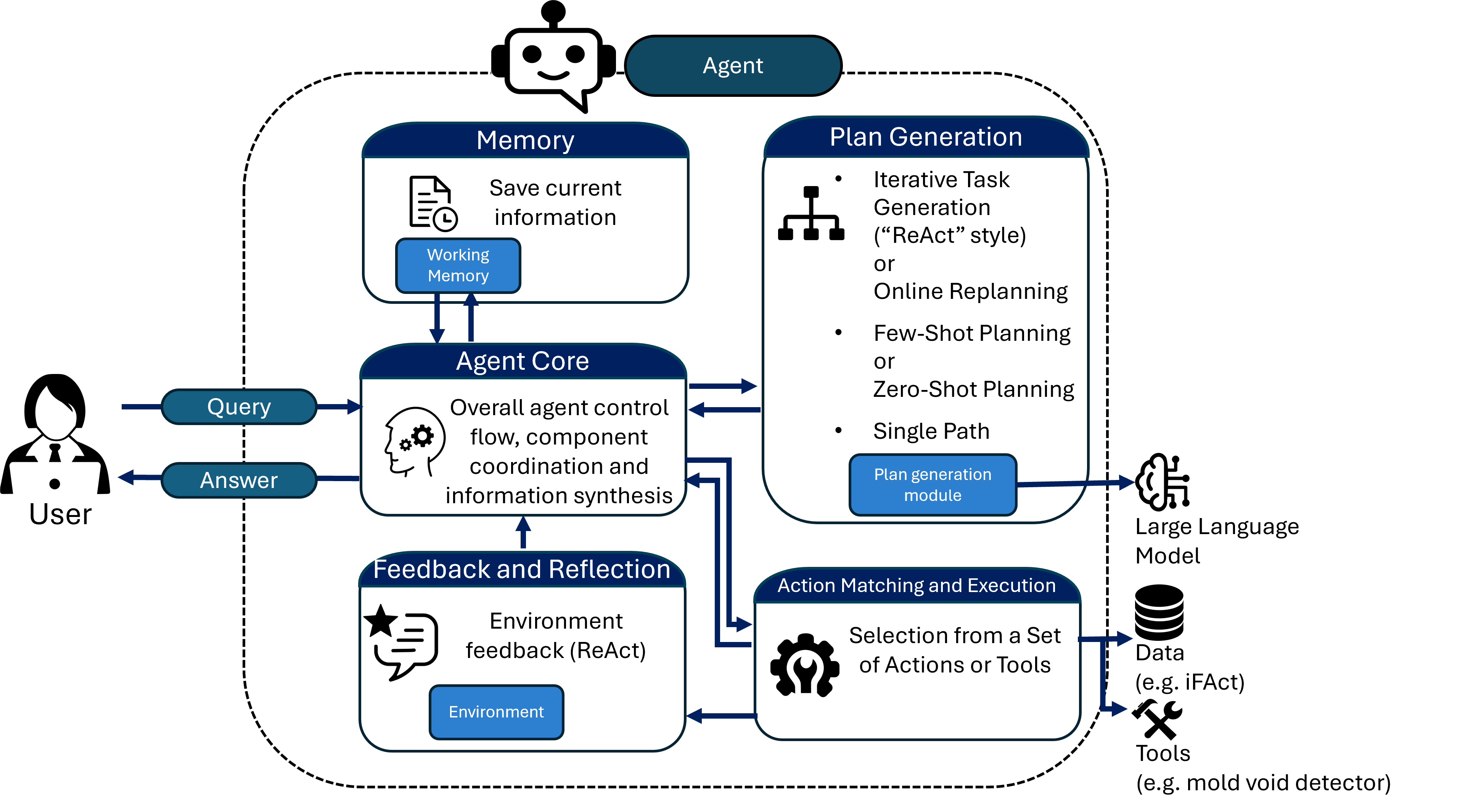}
	\caption[]{ReAct-based LPA architecture for FA applications}
	\label{fig:agentdesigndraft}
\end{figure*}

Figure \ref{fig:agentdesigndraft} shows an abstract agent design with the selected components. 
The system employs two planning approaches for FA applications. The first approach, online replanning, generates a complete plan and adapts it dynamically based on feedback after each step. This approach aims to ensure consistency and avoid endless loops by incorporating criteria for valid answers.
The second approach, iterative task generation with a ReAct agent, interleaves reasoning and action execution, utilizing environment feedback to refine subsequent actions. This method, inspired by factual question answering. 
ReAct \cite{DBLP:conf/iclr/YaoZYDSN023} is one of the most influential and widely adopted planning methods. It employs a thought-action-observation framework, which defines an iterative process involving action selection, execution, retrieval of observations, and the use of these observations to plan subsequent actions. The overall workflow is illustrated in Figure \ref{fig:reactflow}. The ReAct approach has demonstrated superior performance across various tasks compared to standalone LLMs, as it effectively integrates real-time information retrieval through web interaction tools with advanced reasoning capabilities.

LPA actions are usually implemented as a set of tools that a planner can use to interact with the environment, like APIs for retrieving FA data or AI models for specific FA tasks. Tool descriptions must be manually curated for effective integration into prompts, following practices from \cite{shen2024hugginggpt}. Feedback from tools is processed to ensure compatibility with the agent's reasoning context.

%
%



%
\section[LPA Technical Design and Implementation]{Technical Design and Implementation of the LLM-Based Planning Agent} \label{section:architecture}

This section outlines the design and implementation of the Proof-of-Concept LPA, detailing the core design decisions and system architecture. Given the novelty of LPA systems and the lack of industrial examples, a flexible and iterative approach was chosen, especially for prompt design, which required extensive prototyping. The final application was refactored into a modular, object-oriented Python program. Each tool is encapsulated as a class; prompts, logs, and domain-specific data are stored in structured directories. A global settings file manages model selection and paths. Prompts are stored as editable text files, dynamically filled with runtime values and FA-specific knowledge. Logging is implemented for system, user-facing, and evaluation outputs to ensure full traceability. The application is built using the LangChain framework, chosen for its flexible support of agent-based architectures and modular design.


\subsection{Agent Initialization}
\label{subsection:AgentInitialisation}

Agent initialization begins with model selection via the global configuration. Models used in our implementation include Mixtral 8×7B (32k) \cite{mixtral2023}, LLaMA3–70B (8.2k), and LLaMA3–8B (8k) \cite{llama3meta2024}. A $\mathit{temperature}=0$ is used to ensure deterministic responses of an LLM.

\textbf{Tool integration} follows a structured three-step process: (1) the tool logic is encapsulated in a callable function, (2) the tool is wrapped as a LangChain tool with a name and a natural language description (used by the LLM for selection), and (3) it is added to the agent’s tool list. All tools accept a single text input. To ensure robustness, each tool class includes a method to verify availability by checking connectivity and authorization, so that only operational tools are included at runtime. This prevents execution errors and avoids confusion in tool selection. Once tools and models are initialized, the agent can be created and launched using LangChain's \texttt{AgentExecutor}, which orchestrates the reasoning loop and tool invocation.

\begin{python}[caption={Definition of \textsf{ElasticSearch} as LangChain tool}]
query_elastic_failure_analysis_jobs_database _tool = Tool(
name="query elastic failure analysis jobs database",
func=elasticsearch_tool.query_elastic_failure_analysis_jobs _database,
description="Useful if you need to retrieve information about failure analysis jobs, specific devices/products and tasks and findings of previous analyses. This tool allows searching for specific information in the elastic database containing information about all failure analysis jobs. It returns search results of the database query. Input is a text with a well formulated specific question or search instruction specifying what information to search for."
)
\end{python}

\textbf{Prompt handling} uses modular templates with curly brackets for runtime values (e.g., \texttt{{input}}) and uppercase markers (e.g., \texttt{ELASTICFIELDS}) for previous domain-specific insertions. Templates are loaded and formatted at runtime using LangChain utilities, supporting reuse and maintainability.

\textbf{Intent Classification} optionally precedes execution to tailor prompts based on user queries. The LLM assigns one of five predefined categories: (0) root cause or next analysis step suggestion, (1) general failure modes, (2) similar job identification, (3) related general information, or (4) no classification possible. Tested on 20 labeled queries, the classifier performed well overall but showed some ambiguity-related errors.

\subsection{ReAct Agent Implementation}

The ReAct agent is implemented using LangChain's ReAct prompting framework, extended for domain-specific adaptation and increased robustness in production-like environments. Two general prompt versions are supported: the standard LangChain ReAct prompt and a custom variant based on intent classification. Only zero-shot prompting is used in this work. Early tests with few-shot prompting, which included sample ReAct loops, resulted in degraded output quality. LLMs often copied irrelevant example details (e.g. job IDs), corrupting the results. In contrast, zero-shot prompts with clear instructions and domain context proved more stable and effective.

The ReAct prompt uses placeholders for the user input, tool names and tool descriptions. These are dynamically populated with the user's query and  the list of available tools (name, input type, description). The prompt follows a structured, iterative format of \texttt{Question, Thought, Action, Action Input, Observation}. Each interaction is stored in an agent scratchpad, which accumulates the full reasoning history to guide subsequent steps. A final response is generated only when the LLM decides that sufficient information has been gathered.

The agent is executed via a custom wrapper that constructs the ReAct chain using the selected prompt, LLM and tool list. LangChain's \texttt{AgentExecutor} manages the Thought–Action–Observation loop until a final response is produced. Parameters such as timeout, maximum iterations, and parsing error behavior are configurable. A callback handler records each step for later analysis. The complete control flow is illustrated in the sequence diagram (Appendix~\ref{fig:SequenceReAct}).

\begin{lstlisting}[language={prompt},caption={Excerpt from an exemplary agent scratchpad with inserted information}] 
Thought: I need to find similar failure analysis jobs that have a short between a signal and a power domain, and ....  
Action: query elastic failure analysis jobs database 
Action Input: ""short between signal and power domain"" AND ""bent down lead"" AND ""lead frame""
Observation: [""{'Projectnr': '...', 'Package': '.', ... 'No delamination seen ...
\end{lstlisting}

Model-specific issues were observed during testing: while Mixtral handled the standard ReAct prompt correctly, LLaMA3 models failed to stop at the ``Observation'' keyword, probably due to tokenization differences. This broke the loop logic. A variant using ``Response'' was introduced, along with a mechanism to conditionally switch stop sequences depending on the model.

To handle long execution loops, the agent logic has been modified to avoid context window overflow. By default, LangChain appends the full scratchpad at each step, which may exceed the context window. To prevent this, the scratchpad is dynamically trimmed to half the available context window. This required overriding LangChain's default agent creation function to ensure stable operation.

\subsection{Online Replanning Agent Implementation}

The online replanning agent was implemented from scratch using LangChain components and defined in a dedicated class. Unlike the ReAct agent's reactive loop, this architecture follows a structured plan–act–replan cycle, with separate LLM prompt chains for each stage. Following optional intent classification, the system uses a planning prompt to generate an initial step-by-step plan via a dedicated \texttt{planner} chain. Both zero-shot and few-shot styles were tested. Each step is then executed by a simplified ReAct-style \texttt{execution agent}, which receives the current task and prior observations. The result is appended to the execution history.

After the completed action execution, a \texttt{replanner} chain revises remaining steps based on newly acquired observations. Past steps are preserved and excluded from modification, achieved through careful prompt engineering. This loop continues until the plan is completed or a 15-step limit is reached. The current step in the plan is explicitly maintained across iterations. To reduce LLM inconsistencies, planner and replanner outputs are formatted as enumerated lists. Due to the absence of structured output support at implementation time, step parsing is handled via regular expressions. Although effective, this remains brittle and should be replaced with structured output parsing in future versions.

A key challenge was ensuring that only future steps are altered during replanning, requiring precise prompt instructions. Another recurring issue was the generation of overly detailed, human-centric steps (e.g., manual electrical testing, documentation), which are unsuitable for an agent's autonomous execution. As in the ReAct agent, accumulated observations are truncated to half the context window to avoid overflow. After completion, a final \texttt{evaluator} prompt synthesizes the answer from the latest observation, prior steps, and an output directive tailored to the use case from intent classification. The full agent loop is illustrated in the sequence diagram (Appendix~\ref{fig:SequenceOnlineReplanner}).


\subsection{Transparency and Robustness Additions}
\label{subsection:transparency}

To ensure production readiness, the agent includes measures for robustness, transparency, and short-term memory support.

\textbf{Error handling} covers all critical failure points, including file I/O, LLM API access (e.g., outages), and external tool responses. Exceptions are logged and handled gracefully, either continuing with reduced functionality or halting with a controlled message, depending on severity.

\textbf{Logging} operates on two levels. A technical log records system events, warnings, and errors in a timestamped rotating CSV file. A parallel user-facing log records intermediate observations, including step, type, content, and references, structured for display and downstream analysis.

\textbf{Short-term memory} enables later reasoning steps to access earlier tool outputs. A dedicated memory class stores observations as structured CSV entries, retrievable by type. To remain within LLM context limits, retrieved entries are trimmed to fit half the context window.

These mechanisms enhance the system's stability, traceability, and suitability for integration into real-world engineering workflows.


%

\subsection{Technical Design and Implementation of the LLM-Based Planning Agent's Actions}

Agent actions are realized through external tools, typically invoked via internal APIs to access information or perform computations beyond the agent LLM's pretrained scope. Tool selection must balance functionality and context length, while ensuring data security through internal services only. Three categories were integrated: knowledge retrieval tools, an ML-based analysis tool, and explicit reasoning tools.



\paragraph{FA Knowledge Retrieval from ElasticSearch}

ElasticSearch is used to query Infineon’s internal FA job data, providing structured access to historical iFAct job records and documents. Two tool functions were implemented: one retrieves jobs by JobID, the other enables general querying from natural language. Queries are expressed in ElasticSearch’s Query DSL (a JSON-based domain language) and executed using the Python client via \pythoninline{es.search(index='...', body=query)}.

For general queries, a single LLM-driven tool was implemented instead of separate tools per field. This enables complex, multi-field queries and avoids agent prompt bloat. The LLM receives a task prompt, a list of valid fields, and field explanations to construct DSL queries. The output is parsed from unstructured text into JSON and used for API calls. Returned results are cleaned of metadata and unnecessary strings to save context space. If the DSL syntax is incorrect, a repair mechanism triggers up to two LLM-based retries using the original prompt and error feedback. This design allows flexible, contextual querying while keeping prompts compact, a key trade-off in production-grade LPAs.

\begin{lstlisting}[language=json,caption={Example of complex query that was generated within a normal program flow for the question ``What is the failure mode `Fused Wire' related to?''}]
"{ "query": {     
	"bool": {
		"must": [{"match": {"ImageLabel": "Fused Wire"}}]}},   
	"size": 5,  
	 "_source": ["Projectnr", "ImageLabel", "JobSummary"] }" 
\end{lstlisting}

\begin{lstlisting}[language={prompt},caption={Example of a generated description explaining the generated query}]
"Description: This query searches for documents where the "ImageLabel" field contains the phrase "Fused Wire". It returns the top 5 results, including the columns 'Projectnr', 'ImageLabel', and 'JobSummary'. The 'JobSummary' column may provide more information about the failure mode and its context."
\end{lstlisting}

\paragraph{General Knowledge Retrieval from Confluence Pages}

Confluence is Infineon’s internal wiki system containing specialist FA knowledge and product data. The tool enables keyword-based search via the Confluence Query Language (CQL) using the Atlassian Python API. For each user query, text and site search return relevant page IDs, from which titles and body content are retrieved. An LLM then evaluates each page's relevance to the original question. Only relevant pages are summarized and returned as observations, reducing context usage while preserving meaningful content. A mode excluding known evaluation-answer pages is available for test scenarios.

\paragraph{Integration of a Retrieval Augmented Generation Pipeline}

A Retrieval-Augmented Generation (RAG) service, developed in a parallel thesis project \cite{10691083}, was also integrated. It enriches LLM responses by injecting relevant content of retrieved, domain-specific documents into the prompt. The service indexes FA-specific documents and is accessed via FastAPI using simple GET requests. RAG was tested both as a standard agent tool and as a preprocessing step to enrich the query prior to agent execution.

\paragraph[ML Action: Image Retrieval and Mold Void Checking]{Machine Learning Action: Image Retrieval and Mold Void Checking}

To demonstrate ML integration into the LPA workflow, a mold void detection model was added as a tool. It analyzes scanning acoustic microscopy (SAM) top-scan images of TDSO-packaged devices, returning bounding boxes and confidence scores. The tool chain includes (1) image retrieval from the iFAct API, (2) pre-processing with a split-image service, and (3) model inference via API. Only SAM images labeled as ``top'' or ``front'' are selected. Inputs are base64-encoded images; outputs are post-processed to summarize voids by confidence range. Since the model is trained only on TDSO packages, a warning is issued for incompatible images. This tool illustrates how ML services can be orchestrated within LLM-agent workflows, extending capabilities beyond text-based retrieval.

\paragraph{Reasoning Actions with Prompting: Abductive and Practical Reasoning}

Two explicit reasoning tools were implemented to complement the agent's implicit reasoning: abductive reasoning (inferring likely root causes from partial observations) and practical reasoning (suggesting next analysis steps). Both are implemented as prompt-based tools and can access prior observations via the agent's short-term memory. Each tool loads a reasoning prompt, optionally appends previous observations, and prompts the LLM to generate structured reasoning outputs. Two prompt variants were tested: (a) simple prompts requesting hypotheses or recommendations, and (b) explanatory prompts guiding the LLM through step-by-step reasoning, including rating hypotheses or assessing action consequences. Both were evaluated in ablation studies.

For abductive reasoning, the LLM receives prior observations and is instructed to generate 1–3 hypotheses for possible failure causes. A structured prompt version additionally guides the LLM to rate each hypothesis from 0 to 10 based on how well it explains the data. For practical reasoning, the LLM receives prior hypotheses and findings and is asked to recommend concrete, goal-driven next analysis steps. A prompt provides additional structured information including general FA workflows and harmonized task names. This modular prompting approach enables the LPA to perform explicit, interpretable reasoning steps, supporting transparency, testability, and integration of domain knowledge into the agent's decision-making.


%
\section{Evaluation} \label{section:evaluation}

To assess the performance and robustness of the implemented LLM-based planning agent (LPA), we conducted extensive ablation studies across different configurations. The goals were to evaluate (1) technical robustness, (2) answer quality compared to gold standards, and (3) the impact of various architectural and tool choices.

\subsection{Experiment Setup and Execution}

A total of 15 representative FA-related questions were used. All configurations are shown in Table \ref{tab:evaluation_componentconfigurations}. 
While the full combinatorial space would result in 1,296 configurations (19,440 runs), only scientifically interesting combinations were selected. Configurations without core retrieval tools like ElasticSearch or Confluence were excluded, as their absence would clearly degrade performance. Answer quality was evaluated automatically using a Mixtral 8x7B model prompted to score the generated output against gold standard answers on a 1–5 Likert scale. All test executions and evaluations were fully automated. Each run was logged in an evaluation CSV, including configuration details, tool usage, the test question, the gold answer, the evaluated score, and the score reason. Invalid runs (e.g., tool or model failures) were rerun. The final logs were structured into tables and pivot charts for grouped analysis across configurations. Over 2000  valid runs were were performed in total.

\begin{table}[ht]
	\begin{tabular}{p{0.23\linewidth} | p{0.67\linewidth}}
		\hline
		\textbf{Component} & \textbf{Tested Variants}\\
		\hline
		\hline
		Model & Mixtral 8x7B, Llama3 70B, Llama3 8B\\
		\hline
		Agent & ReAct, online replanning zero-shot, online replanning few-shot\\
		\hline
		Intent classification & with intent classification, without intent classification\\
		\hline
		Short-term memory & with memory, without memory\\
		\hline
		Abductive reasoning & simple abductive reasoning prompt, explanatory abductive reasoning prompt, no distinct abductive reasoning tool\\
		\hline
		Practical reasoning & simple practical reasoning prompt, explanatory practical reasoning prompt, no distinct practical reasoning tool\\
		\hline
		Retrieval augmented generation tool & with RAG tool, without RAG tool \\
		\hline
		Retrieval augmented generation preprocessing & with RAG preprocessing, without RAG preprocessing\\
		\hline
		\hline
\end{tabular}
\caption{Overview over tested component configurations}
\label{tab:evaluation_componentconfigurations}
\end{table}

\subsection{Experiment Results}

Key findings from the evaluation are summarized below. All scores represent average values over the 15 representative FA-related questions.

\begin{itemize}
    \item \textbf{ReAct outperformed online replanning} in both runtime and answer quality. ReAct averaged 3.5 minutes per run vs. 14 minutes for online replanning. The best ReAct setup without RAG but with intent reached 3.25 (avg. 2.63), compared to 3.07 (avg. 2.83) for the best online replanning configuration.
    
    \item \textbf{Mixtral 8x7B outperformed Llama3 70B} in ReAct setups; Llama3 70B performed slightly better in online replanning.
    
    \item \textbf{Intent classification improved performance} and was present in all top configurations.
    
    \item \textbf{Short-term memory and reasoning tools} (abductive, practical) showed no consistent benefit.
    
    \item \textbf{RAG improved answer quality}, especially as preprocessing. The best configuration reached 3.6 (avg. 3.14), but added ~3 minutes runtime.
    
    \item \textbf{Llama3 8B underperformed} with a best score of 2.47 (avg. 2.13) and was excluded from RAG tests.
\end{itemize}

The best-performing configuration across all experiments, achieving a best average answer quality of \textbf{3.6}, was:\\
\textbf{ReAct agent}, \textbf{Mixtral 8x7B}, \textbf{intent classification}, \textbf{short-term memory}, \textbf{no abductive reasoning}, \textbf{simple practical reasoning}, and \textbf{RAG integration as preprocessing}.

Overall, answer quality is mixed but promising for a PoC, with many partially useful answers indicating feasibility. Considering technical robustness, all configurations executed reliably. Failures due to service unavailability or timeouts were detected by log checking and resolved. The system consistently produced answers, or correctly indicated when time or iteration limits were exceeded. This confirms the implementation is robust enough for deployment, practical use and user-facing evaluation.

\subsection{Experiment Discussion}

The evaluation confirms the technical feasibility of LPA-based support for failure analysis. While average answer quality was moderate (2–3), top configurations reached up to 3.6, demonstrating that useful responses can be generated. However, answer quality remains sensitive to both prompt formulation and the availability of relevant information.

Although a large number of runs were conducted, the significance of the results should be interpreted with caution. All prompt-based components are highly dependent on the exact wording and structure of their prompts; thus, the results reflect the performance of specific prompt implementations rather than the general effectiveness of the component types. Furthermore, in the initial experiments, question characteristics, such as clarity and the retrievability of relevant information, had a greater impact on answer quality than configuration changes. This highlights the importance of better knowledge integration and more comprehensive test coverage.

The LLM-based evaluation method proved scalable for large test sets but remains inherently brittle and subjective. Future work should incorporate user studies and well-defined ground truth data to validate practical utility. Developing a benchmark dataset with diverse, real-world FA questions and expert-validated answers would significantly enhance the robustness of future evaluations.


%

\section{User Interface Creation and Cloud Deplyoment} \label{section:interface}

To make the LPA application accessible to engineers, a simple user interface was developed using \texttt{Streamlit}, a lightweight Python framework for web-based applications. The interface enables users to submit questions, monitor agent progress, and inspect results. 

\begin{figure}[h]
	\centering
	\fbox{\includegraphics[width=1.0\linewidth]{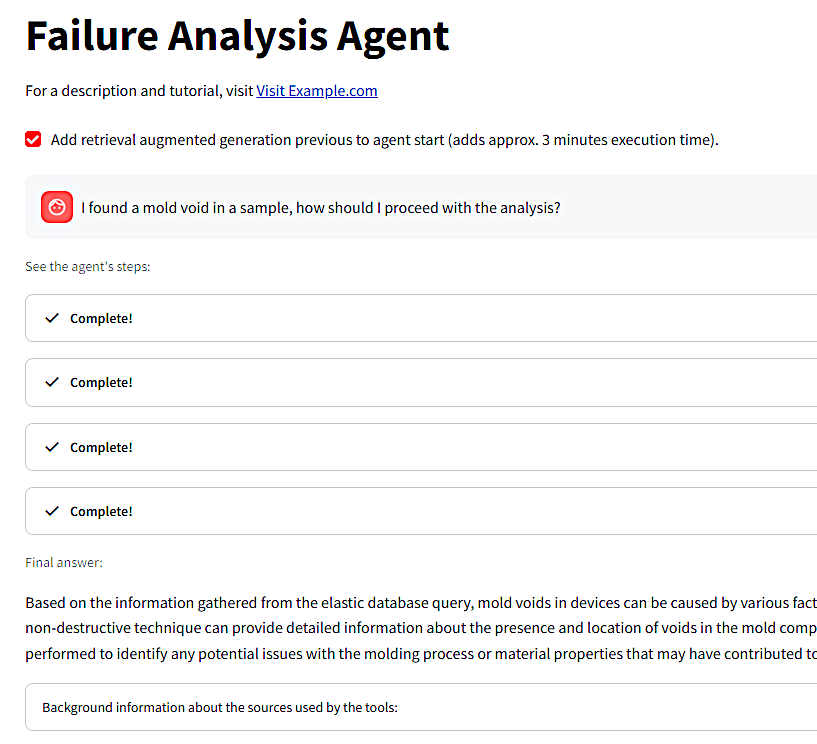}}
	\caption[]{Screenshot of the Failure Analysis Agent application: UI after finishing the processing, displaying the whole trajectory including the answer..}
	\label{fig:faagenttrajectory1}
\end{figure}

Upon launch, users can optionally activate RAG preprocessing via checkbox, trading off improved answer quality against increased runtime. A text field accepts natural language questions; execution begins via a button press. During runtime, users are shown intermediate steps from the agent's reasoning trajectory using LangChain callback handlers. Each action/observation pair can be expanded for inspection. After execution, the full reasoning trajectory and the final answer are displayed. Below, a structured table presents intermediate outputs for each tool. Informative error messages are shown for tool or model failures (e.g., unavailable LLM or disconnected tools). If critical LLM components fail, execution halts with a message. Tool unavailability is reported, and partial processing continues. A built-in feedback mechanism allows users to rate answers (1–5) and leave optional textual feedback. All entries (question, answer, config, tool status, rating, comment, timestamp) are logged anonymously for evaluation. This minimal but functional interface enables hands-on experimentation and evaluation in realistic usage scenarios and prepares the groundwork for future usability studies. Further screenshots are included in the appendix. 



To demonstrate productive applicability, the LPA application was deployed on Infineon's internal cloud platform. 
Deployment was conducted via an internal \textsf{GitLab} repository. Secrets (e.g., credentials) and configuration paths were defined as environment variables and accessed in the application via \texttt{os.environ}. To enable concurrent agent sessions, UUIDs were appended to filenames to prevent file conflicts. Logging and data files were transferred via OpenShift CLI using \texttt{oc cp} commands. This deployment served both as a working prototype for users and as a showcase for internal stakeholders, proving the LPA's technical deployability within enterprise IT environments.


%
\section{Conclusion and Future Work} \label{section:conclusion}

This work demonstrated the feasibility of applying Large Language Model-based Planning Agents (LPAs) to support semiconductor FA. A fully functional PoC agent was designed, implemented, and deployed within Infineon's IT infrastructure, integrating LLM reasoning, external tool use, and autonomous task decomposition. The agent processed complex technical queries and delivered structured, human-readable responses.

Developed through a structured, waterfall-aligned methodology, the system addressed FA-specific requirements and integrated productive tools, including databases and proprietary services. Evaluation showed moderate average answer quality, with the best configuration using the ReAct agent, Mixtral 8x7B, intent classification, and RAG preprocessing. The agent proved technically robust, but performance was sensitive to prompt formulation and available knowledge.

Despite the novelty of the LPA approach, integration into existing infrastructure succeeded. Challenges such as prompt brittleness and tool interoperability were mitigated through modular design, iterative prototyping, meticulous error handling, and externalized configuration. The deployed application includes a user interface and logging mechanisms to support practical usage and transparency.

Overall, this PoC lays the groundwork for applying LPAs in FA and similarly complex domains, enabling not only advanced information retrieval, but also automation of high-level reasoning tasks in knowledge-intensive environments.

\paragraph{Future Extensions and Scalability}
Future development should expand the agent’s capabilities by integrating more tools and implementing additional reasoning actions. A key direction is incorporating domain-specific backgorund knowledge, either through prompt design or fine-tuning. Multi-agent setups, where LLMs collaborate or critique each other, could enhance performance but raise resource concerns. Human-in-the-loop mechanisms may further improve reliability. For full scalability, secure user management and concurrent operation must be addressed.

\paragraph{Added Focus on User Experience and Evaluation}
Usability testing with FA engineers is essential to refine the interface and interaction flow. Observations of real-world use and structured evaluations should inform system adaptation. Additionally, creating a benchmark dataset with realistic FA queries and expert answers would enable consistent, objective performance assessment.

\paragraph{Expected Advances in LLM-Based Planning Agents}
As LLMs evolve, newer models with built-in planning capabilities or fine-tuned on FA data may significantly boost agent performance. Ongoing research into agent architectures, reasoning strategies, and tool orchestration will be key. Sharing frameworks and evaluation methods across domains can accelerate maturity and adoption of LPA systems in industrial contexts.


%
\bibliographystyle{ieeetr}
\bibliography{bibliography}

\appendix

\onecolumn
\section{Appendix}

\subsection{Selected Prompt Templates}
\label{appendix:prompts}

\begin{lstlisting}[language={prompt},caption={Excerpt from a prompt template used for \textsf{ElasticSearch} queries}]
You previously had the following task: 
Given an input question, create a syntactically correct Elasticsearch query ...
...
Only use column names from the following list: (ELASTICFIELDS). 
ELASTICEXPLANATION 
TASKLIST
...
Question: {input}

Now this is your task:
You created the following Elasticsearch query in the previous task: {query}
This query is not a syntactically correct Elasticsearch query.
...
\end{lstlisting}

\begin{lstlisting}[language={prompt},caption={Adapted zero-shot ReAct prompt template}]
Answer the following questions as best you can. You have access to the following tools:
	
{tools}
	
Use the following format:
	
Question: the input question you must answer
Thought: you should always think about what to do
Action: the action to take, should be one of [{tool_names}]
Action Input: the input to the action
Observation: the result of the action
... (this Thought/Action/Action Input/Observation can repeat N times)
Thought: I have gathered detailed information to answer the question considering the intent
Final Answer: the summarized findings and final answer to the original input question and include also the most important and most probable hypotheses you found, with a short explanation, and the most suitable actions to recommend, with a short explanation.
	
Begin!
	
Question:{input} 
Intent: I want to find out about the possible root cause of this failure and receive suggestions for the next action to do in the analysis. Therefore I need to gather information from available sources first. After that I have to do abductive reasoning to formulate hypotheses about the root cause. If I am not sure about the hypotheses, I should gather more information. If the user may want next step or method suggestions, I then have to do practical reasoning.
Thought:{agent_scratchpad}
\end{lstlisting}

\begin{lstlisting}[language={prompt},caption={Zero-shot OnlineReplanning planning prompt template, inspired by \cite{LangGraphPlanAndExecute}}]
For the given question, come up with a step by step plan.
This plan should involve individual questions, that if answered correctly one after another will yield the correct answer. Do not add any superfluous steps.
The result of the final step should be the final answer. Make sure that each step has all the information needed. Do not skip steps.
Output the plan in a list format like this: [1....,\\n 2. ...,\\n3...,\\n ...]

User question: {input}
\end{lstlisting}

\begin{lstlisting}[language={prompt},caption={Prompt template for replanning in OnlineReplanning}]
For the given question, you came up with a step by step plan.
This plan involves individual questions, that if answered correctly one after another will yield the correct overall answer. The result of the final step should be the final answer. 
	
Check if, with the learnings from the past steps, the plan is still valid to answer the question or if the future steps to be adapted. You can change existing future steps, add new future steps or delete unnecessary future steps. Do only include new future steps that can be solved via information processing, including finding information, summarizing information, reasoning and similar actions. Include the relevant technical terms in the steps.  Remember your main goal is to find information to help the user answer their question. You must not include steps that need practical analysis, investigation or in general conducting or performing analysis actions on a real device or in the laboratory.
	
If you add future steps add only absolutely necessary steps. Make the plan as short as possible, with as few steps as possible, between 5 and maximum 10 steps overall. If you think you can already answer the question based on the previous results from the past steps, remove all future steps from the plan.
	
The user question was this:
{input}
	
Your original plan was this:
{plan}
	
You have received these past results:
{past_steps}
	
Next step: {next_step}
	
Update your plan accordingly. The overall plan must have less than 11 steps. Update only steps not done yet. Just copy the previous steps before the next step from the original plan into the new plan. Do not change the previous steps. Do not add anything to the previous steps. Each of the previous steps is only one sentence with only the step. Only output all the plan steps and no additional information. 
	
Output the plan in a list format like this: [1. ...,\n 2. ...,\n 3. ...,\n ...]
\end{lstlisting}

\begin{lstlisting}[language={prompt},caption={Excerpt from the prompt template for synthesizing the final answer of the OnlineReplanning agent}]
For the given question, you came up with a step by step plan.
...
The user question was this:
{input}

This was your last plan step and the results:
{last_past_step}

These were your previous plan steps and the results:
{previous_past_steps}
...
Summarize the information to create a conclusion for answering the question. {intent_out}
...
\end{lstlisting}

\begin{lstlisting}[language={prompt},caption={Prompt template for evaluating the relevance of a retrieved \textsf{Confluence} page}]
Your task is to evaluate the relevance of the content text of a html page to answer the complete ort parts of a given question. 
The question is asked by failure analysis experts and they expect relevant information.
Answer always only in the following format: '[<number>,<reason>]' where <number> is 0 if the document is not relevant and <number> is 1 if the document is relevant. <Reason> is the reason why you think the document is either relevant or not relevant. Be concise and do not output something else.
This is the question: {question}
This is the title of the text: {title}
This is the text: {text}
Remember you must use the answer format '[<number>,<reason>]'
\end{lstlisting}

\begin{lstlisting}[language={prompt},caption={Prompt template with a simple explanation of abdutive reasoning}]
Perform abductive reasoning and find the most likely explanation for the given observations. The objective is to identify possible root causes of observed failures in semiconductor devices given a set of observations like observed failures and existing analysis results.
Generate three hypotheses and output them in a list. 
FAULTS_INFORMATION_STRING
Observations: {observations}
Question: {question}
\end{lstlisting}

\begin{lstlisting}[language={prompt},caption={Prompt template with a comprehensive explanation of abductive reasoning}]
Your task is to perform abductive reasoning, meaning to find the most likely explanation for given observations. The objective is to identify possible root causes of observed failures in semiconductor devices given a set of observations like observed failures and existing analysis results.
Perform the following steps:
1. Identify the set of relevant observations. Observations relate to observations made in analyses of devices and can be observed failures, measurements and general findings from analyses. Specify only the concrete observations. 
2. Generate hypotheses that can best explain the set of observations. The hypotheses must explain why the observations occurred and can be potential root causes or potential failures. Generate at least 1 and at most 3 hypotheses.
3. For every generated hypothesis, rate how well the hypothesis explains the observations. Rate by assigning an integer value between 0 and 10 to the hypothesis, where 0 means the hypothesis is absolutely not likely to explain the observations and 10 means the hypothesis most probably explains the observations.
FAULTS_INFORMATION_STRING
Observations: {observations}
Question: {question}
\end{lstlisting}

\begin{lstlisting}[language={prompt},caption={Exemplary abductive reasoning tool output when using the simple prompt template}]
Based on the given observations, here are three possible hypotheses for the root cause of the observed failures in semiconductor devices:  
	1. "PinShort": The short in the device could be caused by a short circuit between pins, which may be due to manufacturing defects, contamination, or damage during assembly. 
	2. "MetallizationPinholes": The short could be caused by pinholes in the metallization layers, which may allow for electrical connections between different layers or components, leading to a short circuit. 
	3. "ParticleShortingPinsLeads": The short may be caused by particles or foreign materials that have bridged the gap between pins or leads, creating a short circuit. This could be due to contamination during manufacturing, handling, or assembly processes.
\end{lstlisting}

\begin{lstlisting}[language={prompt},caption={Exemplary abductive reasoning tool output when using the explanatory prompt template}]
	...
1. Identify the set of relevant observations: 
- Short in a device which is not decapsulated yet 
- Localization of the short in the package 
- IR-LIT - Magnetic Microscopy 
...
2. Generate hypotheses that can best explain the set of observations: 
Hypothesis 1: Solder bridge between pins in the package causing the short
Hypothesis 2: Cracked die or wire causing the short  
3. Rate how well the hypothesis explains the observations: 
Hypothesis 1: Solder bridge between pins in the package causing the short - Explains the localization of the short in the package - Explains the need for IR-LIT, Magnetic Microscopy, ... to identify the short - Rate: 9  
Hypothesis 2: Cracked die or wire causing the short - Explains the short in the device which is not decapsulated yet - Does not directly explain the localization of the short in the package - Explains the need for IR-LIT, ... to identify the short - Rate: 7  
Based on the observations, Hypothesis 1 (solder bridge between pins in the package causing the short) is more likely to explain the observations than Hypothesis 2 (cracked die or wire causing the short)
\end{lstlisting}

\begin{lstlisting}[language={prompt},caption={Explanatory practical reasoning prompt template}]
You are an agent assistant that supports a failure analysis engineer.
Your task is to do practical, goal-based reasoning. Your task is to identify one action or a short sequence of actions for further analysis steps to examine the given hypothesis.
1. Check the given information on hypotheses and previous analyses. 
2. Select the best actions (at least 1, at most 5) that can reach the goal of confirming or rebutting the given hypotheses and align with the analyses. You can select actions from the given failure analysis information or given tasks, but you can also select actions that are described in previous analyses. Select only actions for the immediate next steps. Try to select concrete actions.
3. Also think about the consequences of the actions you propose and if anything could negatively affect the goal of further researching the hypotheses, like destruction.

These are the hypotheses about the root cause and failures and existing previous analyses and results: {hypotheses_and_previous_analyses}

Here is general failure analysis flow information: GENERAL_INFORMATION
These are possible tasks to conduct: TASK_LIST

This is the initial question: {question}

Justify briefly and concisely why you selected the action or actions. If you are not sure about your selected actions also mention that and what additional information you need. Be concise.
\end{lstlisting}

\begin{lstlisting}[language={prompt},caption={Exemplary output when using the practical reasoning explanatory prompt template}]
The following hypotheses were formulated: 1) Corrosion, 2) Moisture-related issues, and 3) Defective bonding.": 
"Based on the given hypotheses and previous analyses, the following actions can be taken to further investigate the root cause of the failures:  
1. Electrical Verification: Validate the reported failure by performing electrical testing on the device, such as Tester Measurement (ATE) and Curve Tracer Measurement (I-V). This will help confirm the intermittent behavior and irregularities observed in the VDDM curve.  
2. Optical Microscopy: Perform external and internal visual inspection using Optical Microscopy (OM) to look for signs of corrosion, moisture-related issues, or defective bonding. This can provide initial visual evidence to support or refute the hypotheses.  
3. X-Ray Radiography: Conduct X-ray imaging to examine the bond wires, solder joints, Cu traces, and lead frames for any visible defects or anomalies that could be related to the failures. This can help isolate the physical failure fault and provide insights into the root cause.  
4. Decapsulation: Perform chemical decapsulation to expose the die and bonding area for further analysis. This will allow for a more detailed examination of the bonding and potential corrosion or moisture-related issues.  
5. (Cross-)subsectioning: Perform cross-subsectioning on the bonding area to examine the bonding quality and look for signs of corrosion or moisture-related issues. This can provide more detailed information on the root cause of the failures.  
These actions align with the general failure analysis flow and the given tools and techniques for fault isolation and defect analysis. They are non-destructive or minimally destructive, allowing for further analysis if needed. However, if the root cause is not identified after these actions, more destructive techniques like SEM, TEM, or FIB may be required.")"
\end{lstlisting}

\FloatBarrier
\newpage
\subsection{ReAct-based LPA Implementation}

\begin{figure*}[h]\begin{center}
	\includegraphics[width=1.0\textwidth]{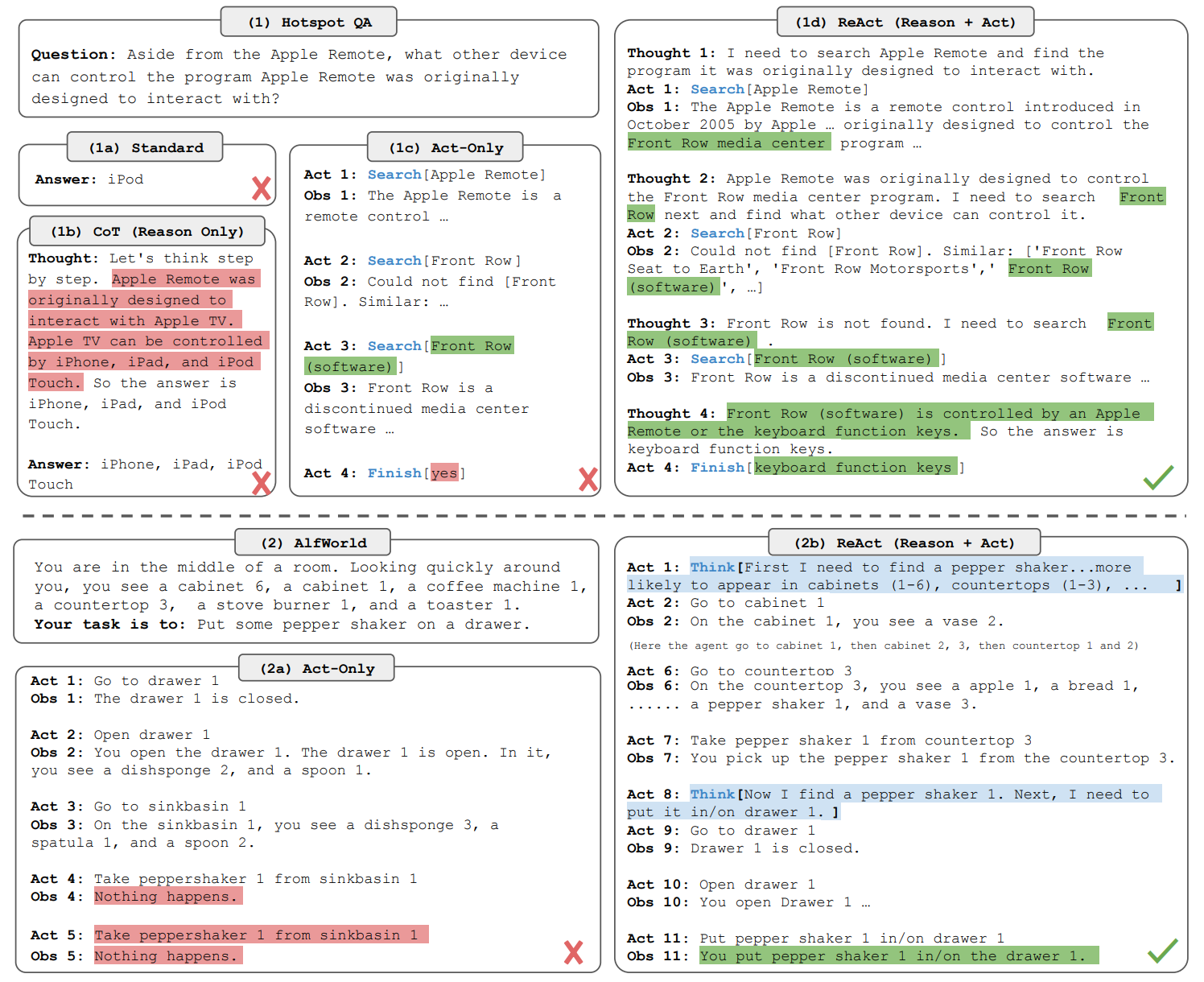}
	\caption{Example of task solving trajectories for different prompting methods 
		on a HotpotQA \cite{yang2018hotpotqa} question requiring up-to-date knowledge retrieval \cite{DBLP:conf/iclr/YaoZYDSN023}}
	\label{fig:reactpromptexample}
\end{center} 
\end{figure*}

\begin{figure*}[h]
	\centering
	\includegraphics[width=1.0\textwidth]{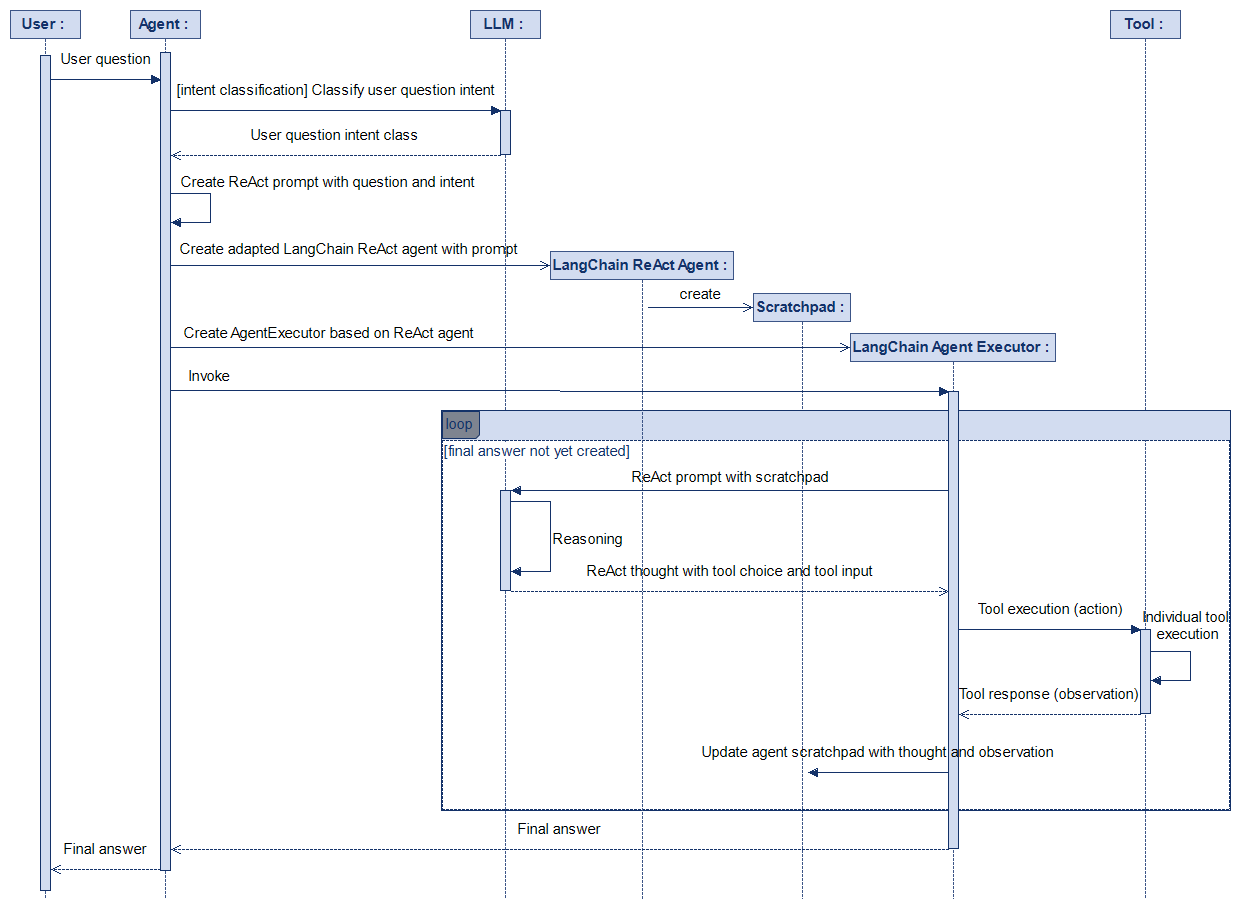}
	\caption[]{ReAct agent program flow sequence diagram (simplified)}
	\label{fig:SequenceReAct}
\end{figure*}

\begin{figure*}[h]
	\centering
	\includegraphics[width=1.0\textwidth]{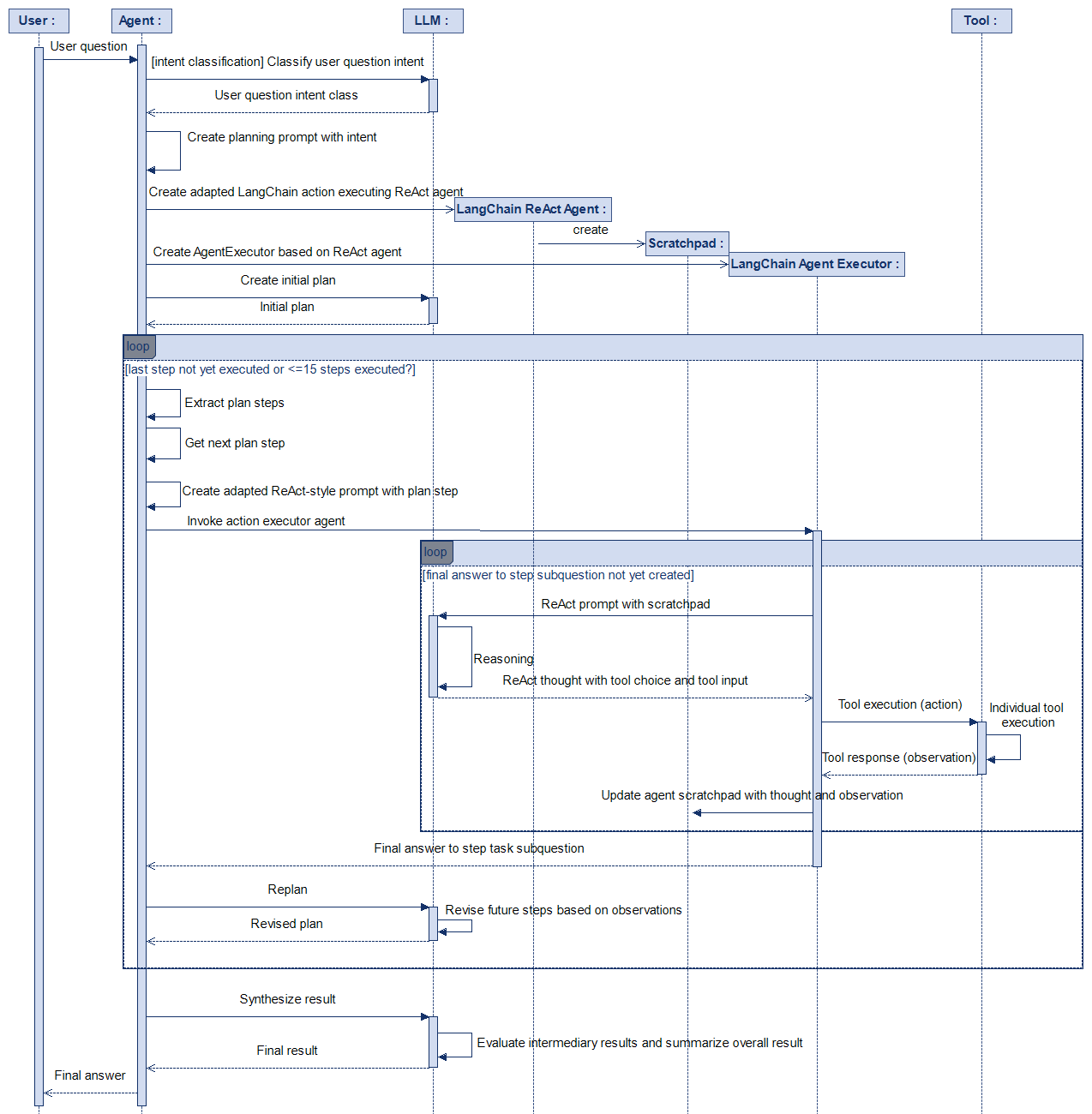}
	\caption[]{Online replanning agent program flow sequence diagram (simplified)}
	\label{fig:SequenceOnlineReplanner}
\end{figure*}


\FloatBarrier

\subsection{Excerpts from Logging}

\begin{figure*}[h]
	\centering
	\includegraphics[width=1.0\textwidth]{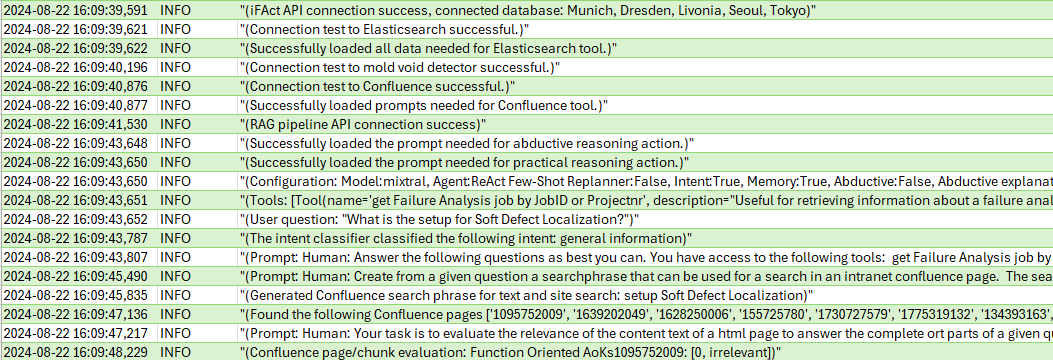}
	\caption[]{Extract from the technical log for events and errors (\lstinline|agent_log|)}
	\label{fig:agentlog}
\end{figure*}

\begin{figure*}[h]
	\centering
	\includegraphics[width=1.0\textwidth]{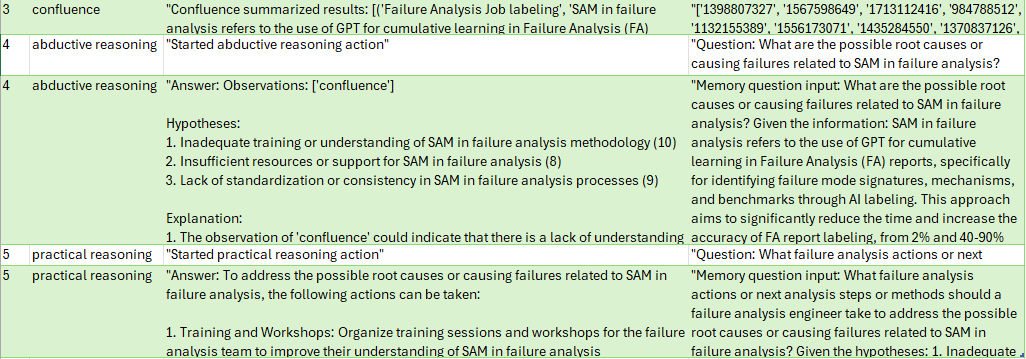}
	\caption[]{Extract from the information log for generating user information (\lstinline|ui_info_log|)}
	\label{fig:infolog}
\end{figure*}

\begin{figure*}[h]
	\centering
	\includegraphics[width=1.0\textwidth]{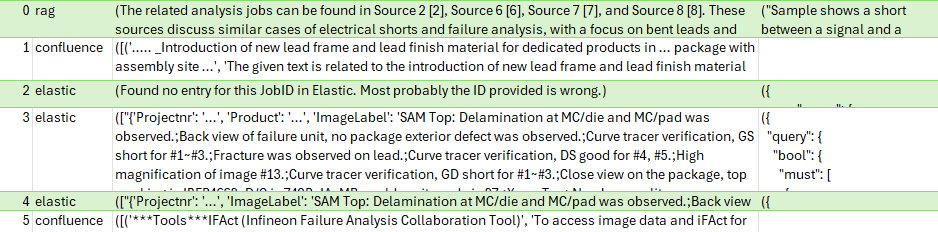}
	\caption[]{Cleaned extract from observation information stored in the agent's short-term memory}
	\label{fig:agentmemory}
\end{figure*}

\begin{figure}[h]
	\centering
	\includegraphics[width=1.0\textwidth]{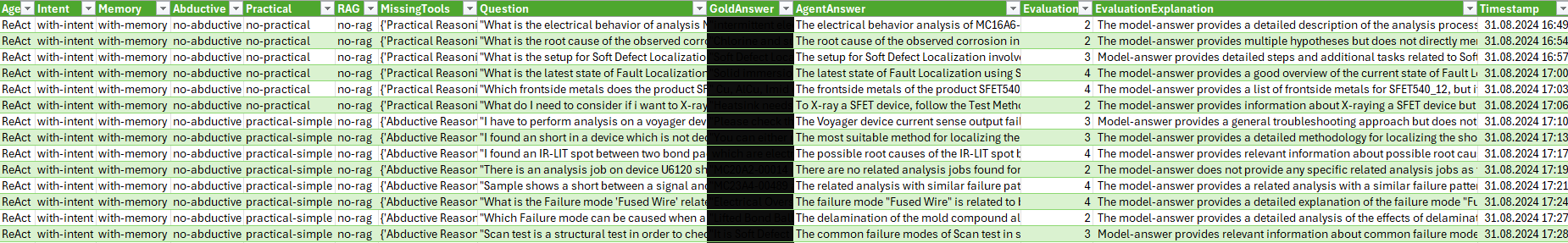}
	\caption[]{Exemplary excerpt from the evaluation log}
	\label{fig:evaluationlogoutput}
\end{figure}

\FloatBarrier
\subsection{Program Screenshots}\label{app:screenshots}

\begin{figure*}[h]
	\centering
	\fbox{\includegraphics[width=1.0\textwidth]{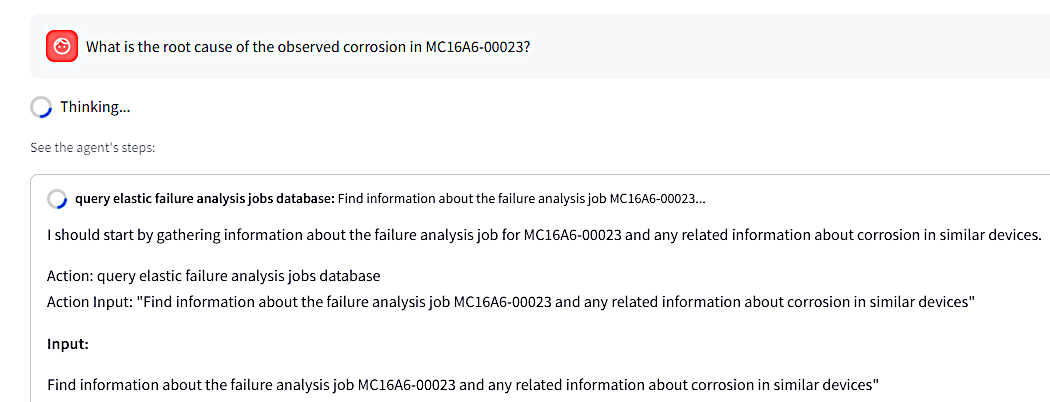}}
	\caption[]{Failure Analysis Agent application: Display of details of intermediate steps}
	\label{fig:faagentmonitoring2}
\end{figure*}

\begin{figure*}[h]
	\centering
	\fbox{\includegraphics[width=0.8\textwidth]{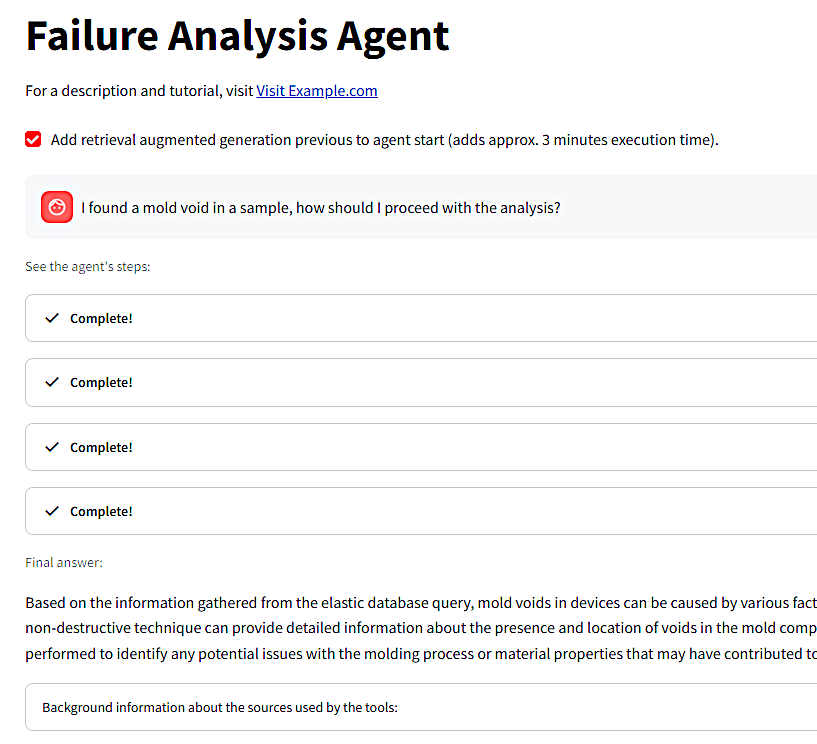}}
	\caption[]{Screenshot of the Failure Analysis Agent application: UI after finishing the processing, displaying the whole trajectory including the answer.}
	\label{fig:faagenttrajectorya1}
\end{figure*}

\begin{figure*}[h]
	\centering
	\fbox{\includegraphics[width=1.0\textwidth]{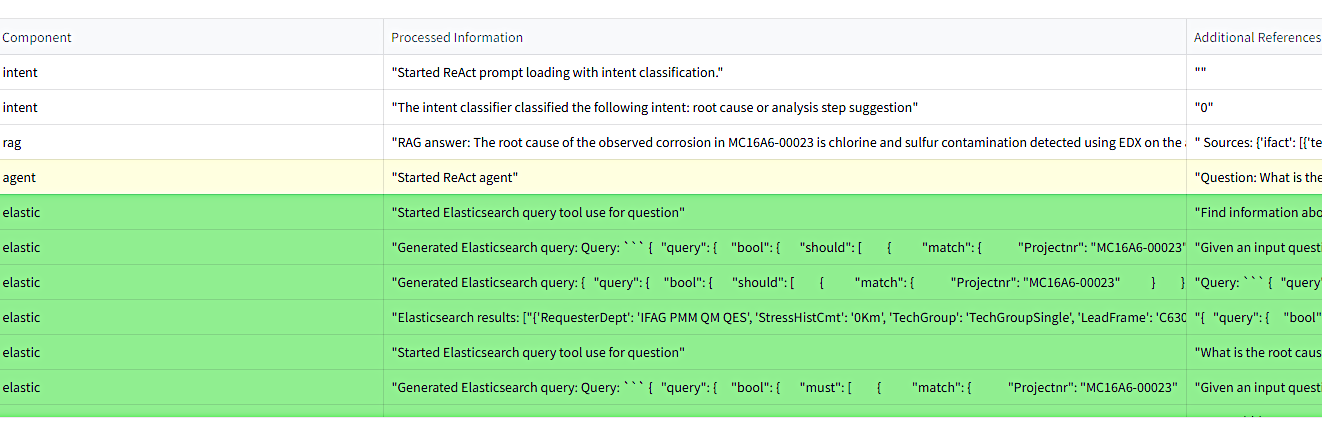}}
	\caption[]{Screenshot of the Failure Analysis Agent application: Additional background information.}
	\label{fig:faagentbackgroundinformation}
\end{figure*}

\begin{figure*}[h]
	\centering
	\fbox{\includegraphics[width=1.0\textwidth]{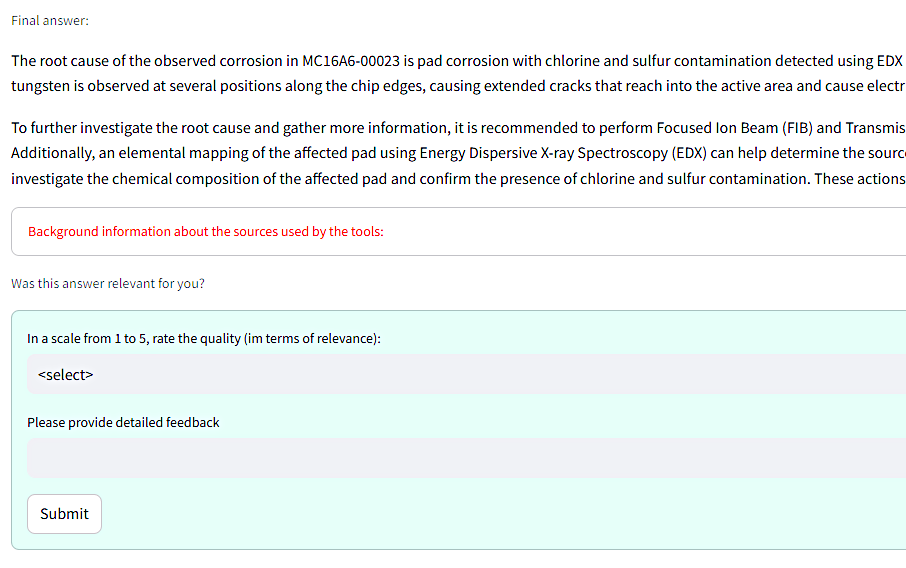}}
	\caption[]{Screenshot of the Failure Analysis Agent application: User evaluation functionality.}
	\label{fig:faagentanswerevaluation}
\end{figure*}

\end{document}